\def\preprintversion{}
\documentclass[letterpaper]{article} 

\ifdefined\preprintversion
\usepackage[draft]{aaai25}
\else
\usepackage[submission]{aaai25}
\fi

\usepackage{aaai25}  
\usepackage{times}  
\usepackage{helvet}  
\usepackage{courier}  
\usepackage[hyphens]{url}  
\usepackage{graphicx} 
\usepackage{natbib}  
\usepackage{caption} 
\usepackage{algorithm}
\usepackage{algorithmic}

\usepackage{newfloat}
\usepackage{listings}
\usepackage{latexsym}
\usepackage{amssymb}
\usepackage{amsmath}
\usepackage{amsthm}
\usepackage{booktabs}
\usepackage{enumitem}
\usepackage{graphicx}
\usepackage{color}
\usepackage{latexsym}
\usepackage{smartdiagram}
\usepackage{float}

\usepackage{tikz} 
\usepackage{listings}
\usepackage{url}

\usepackage{graphicx}
\usepackage{multicol}
\usepackage{array}
\usepackage{todonotes}
\usepackage{float}
\usepackage{subfig}
\usepackage{cuted}
\usepackage{caption}
\usepackage{xcolor}

\newtheorem{example}{Example}

\newtheorem{definition}{Definition}

\newcommand{\myabstract}{{\sf Abst}}
\newcommand{\mysubstring}{{\sf Substrg}}
\newcommand{\myforget}{{\sf Fgt}}
\newcommand{\myatoms}{{\sf Atms}}
\newcommand{\similarity}{{\sf Sim}}
\newcommand{\similaratoms}{{\sf SimilarAtms}}
\newcommand{\extract}{{\sf Ex}}

\DeclareCaptionStyle{ruled}{labelfont=normalfont,labelsep=colon,strut=off} 
\floatstyle{ruled}
\newfloat{listing}{tb}{lst}{}
\floatname{listing}{Listing}
\title{Identification of Entailment and Contradiction Relations\\ for Natural Language Argumentation:\\ 
A Neurosymbolic Approach}
\author {
    Xuyao Feng, Anthony Hunter
}
\affiliations {
    Department of Computer Science, University College London\\
    London, UK\\
    xuyao.feng.20@ucl.ac.uk
}

\usepackage{bibentry}

\begin{document}

\maketitle

\begin{abstract}
Argumentation is an important cognitive activity that involves the scrutiny of arguments and counterarguments.
When given an argument, it is desirable that its premises entail its claim, and that the premises and claim are consistent,
and when given a pair of arguments, it is desirable to check whether one argument supports or attacks another.
If the arguments are in natural language, then we could use methods for natural language inference, also known as recognizing textual entailment (RTE). 
The state-of-the-art RTE methods use machine learning and deep learning, but these are not explainable. 
An alternative approach which we investigate in this paper is to translate text into an abstract meaning  representation (AMR) graph. 
For this we use  a pre-trained AMR parser. 
We then translate the AMR graph into propositional logic and use a SAT solver for automated reasoning. 
In text, often common sense suggests that an entailment (or contradiction) relationship holds between a premise and a claim, but because different wordings are used, this is not identified from their logical representations. 
To address this, we introduce relaxation methods to allow replacement or forgetting of some propositions. 
Our experimental results show our approach performs well on three RTE datasets. 
The benefit of our approach is that we have a logical representation of premises and claims, 
and of the relaxations that are used to identify the entailment and contradiction relations.
\end{abstract}

\section{Introduction}

Humans constantly resort to argumentation when there is conflicting and incomplete information \cite{Atkinson2017}. 
To support real world argumentation, 
humans need to identify entailment and contradiction relations in and between arguments in natural language, such as whether the premises of an argument entails its claim. 

In computational linguistics, methods for natural language inference (NLI), also known as recognizing textual entailment (RTE), are aimed at identifying the semantic relationship between natural language sentences. Given a premise (P) sentence and a claim (C) sentence, the relationship between P and C is classified as entailment, contradiction, or neutral \cite{Padó2015167,PUTRA2023,dagan2006machine}. 

The RTE task is challenging because many natural language sentences assume common sense knowledge for them to be understood. Humans are very capable of reasoning with common sense knowledge. For example, ``The boy jumps." and ``The boy sleeps". These two simple sentences contradict each other as we know a person can not sleep and jump simultaneously. 

The traditional symbolic RTE methods involve {\em logical inference-based methods} \cite{10.3115/1220575.1220654,DBLP:journals/corr/WotzlawC13}, {\em lexical-based methods} \cite{10.1007/11736790_16,jijkoun2005recognizing}, and {\em syntax-based methods} \cite{phdthesis}. The main disadvantage of the symbolic methods is that they are not scalable and require additional information to handle sentences involving common sense knowledge.
More recently RTE methods involve training {\em machine learning} (ML) \cite{10.1007/978-3-030-16660-1_40} or {\em deep learning} (DL) \cite{DBLP:journals/corr/BowmanAPM15,rocktaschel2016reasoning} models that are trained on RTE datasets. As these models need to be trained, they are not so good at general RTE tasks. Also, they are black boxes, meaning the reasoning is not explainable. 
A new development is to {\em combine large language models with symbolic solvers} for logical reasoning tasks similar to RTE \cite{olausson-etal-2023-linc,pan-etal-2023-logic}. Potentially, they offer more robust solutions that are explainable. However, to date, these methods have been evaluated on relatively straightforward datasets where the information is quite explicit (i.e. does not require common sense knowledge). 

For the premises and claim of natural language arguments, we want an RTE approach that does not require training, that is explainable, and that handles some of the problems of common sense knowledge. 
To address these needs, we propose a pipeline that combines a pre-trained large language model that can translate free text into logical formulas, so a form of explanation, and then use neurosymbolic reasoning to overcome some of the challenges of common sense reasoning. The neurosymbolic reasoning uses relaxation methods for identification of both entailment and contradiction. We show that these relaxation methods substantially improve recall, with some loss of precision, for identifying entailment and contradiction on three RTE datasets.

We believe our RTE methods can be combined with argument mining methods (which extract free text premises and claims of arguments and counterarguments \cite{Lawrence2019}). So our methods could be used to translate the outputs of argument mining into propositional logic arguments, and then check relationships within and between them. 

\section{Background}

In this section, we review abstract meaning representation (AMR), and automated reasoning based on SAT solvers, which we incorporate in our pipeline.

\subsection{Abstract meaning representation}

Abstract meaning representation (AMR) is a semantic representation language for representing sentences as rooted, labelled, directed, and acyclic graphs (DAGs). AMR is intended to assign the same AMR graph to similar sentences, even if they are not identically worded. The approach was first introduced by Langkilde and Knight in 1998  \cite{langkilde-knight-1998-generation-exploits} as a derivation from the Penman Sentence Plan Language \cite{kasper-1989-flexible}. In 2013, AMRs re-gained attention due to Banarescu et al.  \cite{banarescu-etal-2013-abstract}, and were introduced into NLP tasks such as machine translation and natural language understanding. The modern (post-2010) AMR draws on predicate senses and semantic roles from the OntoNotes project  \cite{hovy-etal-2006-ontonotes}.

\begin{figure}[t]
\footnotesize
    \[
\begin{array}{ll}
\mbox{\tt (w / want-01}
& \mbox{\tt (w / want-01}\\
\hspace{5mm}     \mbox{\tt    :ARG0 (b / boy)}
& \hspace{5mm} \mbox{\tt :ARG0 (b / boy)}\\
\hspace{5mm}   \mbox{\tt    :ARG1 (g / go-01}
& \hspace{5mm} \mbox{\tt :ARG1 (g / go-01}\\
\hspace{1cm}  \mbox{\tt          :ARG0 b))}
& \hspace{1cm}\mbox{\tt :ARG0 b}\\
&\hspace{1cm}\mbox{\tt  :polarity -))}
    \end{array}
    \]
    \caption{AMR graphs for the sentences ``The boy wants to go." (left) and ``The boy does not want to go." (right).\\}
    \label{f1}
\end{figure}

In AMR, negation is represented via the :polarity relation (see for example, Figure \ref{f1}).
The numbers after the instance name (such as want-01 above) denote a particular OntoNotes or PropBank semantic frame  \cite{kingsbury-palmer-2002-treebank}. These frames have different parameters, but the subject is generally denoted by \texttt{:ARG0} and the object by \texttt{:ARG1}. AMR uses the same structure to represent semantically similar texts by making several simplifying assumptions. AMR cannot represent verb tenses nor distinguish between verbs and nouns. It does not represent articles, quote marks, or the singular and plural.

In our pipeline, we use an IBM Transition AMR parser\footnote{\url{https://github.com/IBM/transition-amr-parser/tree/master}}\cite{DBLP:journals/corr/abs-2112-07790} which is based on the BART large language model  \cite{DBLP:journals/corr/abs-2112-07790}.
This parser translates a sentence of text into an AMR graph.

\subsection{Automated reasoning}

We can easily transform an AMR graph into first-order logic formulas using the Bos algorithm \cite{bos-2016-squib}, which joins all predicates by an implicit conjunction \cite{chanin2023neuro}. 
Each atom in the formula is either monadic or dyadic.
An example of this is shown below, where the AMR for ``The boy does not want to go" from Figure \ref{f1} is converted into a logical formula as follows:
\[
\begin{array}{l}
\tt \exists W ( \exists B(want(W) \wedge ARG0(W, B) \wedge boy(B) \\
\hspace{1cm}   \tt \wedge \hspace{1mm} \neg \exists G (ARG1(W, G) \wedge go(G) \wedge ARG0(G, B))))
\end{array}
\]

Chanin and Hunter \cite{chanin2023neuro} developed an
open-source Python library, based on the Bos algorithm, 
to translate AMR graphs into first-order logic. 
This library is extended in our proposal by simplifying each  first-order logic formula to a propositional logic formula by grounding out the existential quantifiers and replacing variables with new constants. 
Essentially, we are making the assumption that for each existentially quantified variable, there is a specific entity that can represent the quantified variable. This can be viewed as a Skolem constant. 
For the above example, the grounded formula is as follows:
\[
\begin{array}{l}
\tt want(w) \wedge ARG0(w, b) \wedge boy(b) \\
\hspace{2cm}     \tt \wedge \hspace{1mm}  \neg (ARG1(w, g) \wedge go(g) \wedge ARG0(g, b))
\end{array}
\]

We assume the usual definitions for propositional logic.
For automated reasoning, we transform all propositional logic formulas into conjunctive normal form (CNF) using SymPY \cite{10.7717/peerj-cs.103}: CNF is defined as: (1) A literal is either a propositional atom, or the negation of one; (2) A clause is a disjunction of literals; and (3) A formula in conjunctive normal form (CNF) if it is a literal, or a clause, or a conjunction of clauses. 

In our pipeline, we use PySAT \cite{imms-sat18}, which integrates several widely used state-of-the-art SAT solvers as theorem provers to check whether a CNF is consistent. 
Finding an interpretation that fulfils a given Boolean formula is known as the Boolean satisfiability problem (SAT). 
For example, the formula $\tt a \wedge \neg b$ is satisfiable as $\tt a$ = True and $\tt b$ = False resulting in the formula being True. In contrast, $\tt a \wedge \neg a$ is unsatisfiable. 

Proving entailment, i.e $\Phi \vdash \alpha$ where $\Phi$ is a set of formulas, and $\alpha$ is a formula, is equal to showing $(\bigwedge \Phi) \land\neg \alpha$ is inconsistent. To do this, we rewrite $(\bigwedge \Phi) \land\neg \alpha$ into a CNF formula, and then use PySAT to check consistency of this CNF formula.
Similarly, to show  contradiction, i.e. $\Phi\cup\{ \alpha\} \vdash \perp$, is equal to showing $(\bigwedge \Phi) \land \alpha$ is inconsistent, and so we use the CNF with PySAT as before.

\section{Datasets}

We use three datasets in this paper, namely e-SNLI \cite{NEURIPS2018_4c7a167b}, SICK \cite{marelli-etal-2014-semeval} and MultiNLI \cite{N18-1101}, described below.
Since we do not train any component in our pipeline,
we only use the datasets for evaluation, and hence we treat all the data as evaluation data.

The {\bf e-SNLI} dataset is an extension of the Stanford Natural Language Inference dataset  \cite{DBLP:journals/corr/BowmanAPM15} augmented with
a human written natural language explanation for each data item \cite{NEURIPS2018_4c7a167b}. The explanation is meant to explain why the premise entails, or contradicts with, or is neutral with respect to, the claim. 
The dataset consists of 569k items of data.
Each data item consists of a premise, claim, explanation and a label which is one of entailment, contradiction, and neutral. Here is an example where the bold text represents the key concepts (which we explain immediately after the example). 

\begin{example}
    (Premise): A \textbf{man} in an orange vest \textbf{leans over a pickup truck}. (Claim): A man is \textbf{touching a truck}. (Explanation): Man leans over a pickup truck implies that he is touching it. (Label): Entailment.
\end{example}

The premise, claim and label are from the SNLI dataset, and the e-SNLI dataset also include the explanation and the key concepts.
The key concepts are the words that the annotator highlighted in the premises and claim as being essential for the label and the explanation. 
The annotators were asked to write the explanation based on what they had highlighted. 
Also, the annotators had to highlight at least one word in the premise for entailment pairs, 
at least one word in the premise and the claim for contradiction pairs, 
and only words in the claim for the neutral pairs \cite{NEURIPS2018_4c7a167b}. 
In our evaluation, later in the paper, we compare using each {\em premise} alone, with the conjunction of {\em premise and explanation}, to see if the extra information, in the explanation, improves identification of entailment or contradiction relations.

The {\bf Sentences Involving Compositional Knowledge (SICK)} dataset is a dataset for compositional distributional semantics. It covers diverse lexical, syntactic and semantic phenomena \cite{marelli-etal-2014-semeval}. It comprises 
9840 items of data.
Each item consists of a premise, a claim, and a label (entailment, contradiction, or neutral).

The {\bf Multi-Genre Natural Language Inference (MultiNLI)} corpus is a crowd-sourced collection of 433k items. It is modeled on the SNLI  \cite{DBLP:journals/corr/BowmanAPM15} corpus but covers a range of genres of spoken and written text \cite{N18-1101}. Each item consists of a premise, a claim, and a label (entailment, contradiction, or neutral). 

\section{Method}

Our pipeline\ifdefined\preprintversion\footnote{\url{https://github.com/fxy-1117/Neurosymbolic-RTE}}\else\footnote{Link to Github repository hidden for blind reviewing}\fi
which is summarized in Figure \ref{fig:pipeline},
consists of four main components: 
A Text-To-AMR parser; 
An AMR-to-propositional-logic translator; 
A set of methods for relaxation of the logical formulas;
And an automated reasoner based on PySAT. 

\begin{figure}[t]
\begin{center}
\begin{tikzpicture}[->,thick,
arg/.style={draw,text centered, text width=70mm,
shape=rectangle, 
rounded corners=2pt,
fill=gray!20,font=\footnotesize}]
\node[] (a0)  at (0,4) {\footnotesize Sentences for premise and claim};
\node[arg] (a1)  at (0,3.2) {\footnotesize Text-to-AMR Parser};
\node[arg] (a2)  at (0,2.4) {\footnotesize AMR-to-Propositional-Logic Translator};
\node[arg] (a3)  at (0,1.6) {\footnotesize Relaxation Methods};
\node[arg] (a4)  at (0,0.8) {\footnotesize Automated Reasoner based on PySAT};
\node[] (a5)  at (0,0) {\footnotesize Label (entailment, contradiction, and neutral)};
\path (a0)[] edge[] (a1);
\path (a1)[] edge[] (a2);
\path (a2)[] edge[] (a3);
\path (a3)[] edge[] (a4);
\path (a4)[] edge[] (a5);
\end{tikzpicture}
\end{center}
\caption{Our sentence-to-label pipeline. The text input is premises and claim, and the output is a label.}
\label{fig:pipeline}
\end{figure}

\subsection{Entailment methods}
\label{section:entailment}

We introduce two methods to prove entailment between sentences: the {\em propositional entailment method} and the {\em relaxed propositional entailment method}. The former aims to prove entailment with the CNF formula version of the formula obtained from the AMR-to-propositional-logic translator. The latter rewrites the formulas that are obtained from the AMR-to-propositional-logic translator using relaxation methods (which we explain in detail below) and then tries to show entailment with the CNF version of the rewritten formulas. For example, if we have a premise $\tt x_1 \wedge x_2 \wedge x_3$ and a claim $\tt x_1 \wedge x_4$ and we can show that $\tt x_3$ and $\tt x_4$ are very similar concepts, then we can change the claim to $\tt x_1 \wedge x_3$, and then show entailment holds using the CNF versions of these relaxed formulas with PySAT. 

\subsubsection{Propositional Entailment}

An AMR formula is composed from a set of monadic and dyadic atoms together with the $\neg$ and $\wedge$ logical operators and the tautology $\top$ and falsity $\bot$ symbols.
Since each monadic and dyadic atom is ground, we regard an AMR formula as a propositional logic formula.

\begin{definition}
Let ${\cal A}$ be a set of monadic and dyadic atoms. 
The set of {\bf AMR formulas}, denoted ${\cal L}$, is defined inductively:
If $\alpha\in{\cal A}\cup\{\top,\bot\}$, then $\alpha\in{\cal L}$;
If $\alpha,\beta\in{\cal L}$, then $\alpha\wedge\beta\in{\cal L}$;
And if $\alpha\in{\cal L}$, then $\neg\alpha\in{\cal L}$.
\end{definition}

Given a sentence $S$, the AMR parser, together with the AMR-to-propositional-logic translator, gives an AMR formula $\phi$.
We say that $\phi$ {\bf represents} $S$. 

\begin{example}
\label{ex:amrformula}
The AMR formula $\tt \neg (go(g) \wedge ARG0(g,c) \wedge car(c))$
represents the sentence ``the car does not go".
\end{example}

An abstract formula is a propositional formula composed from a set of propositional letters together with the $\neg$ and $\wedge$ logical operators and $\top$ and $\bot$ symbols. We use abstract formulas for automated reasoning with PySAT.

\begin{definition}
Let ${\cal P}$ be a set of propositional letters. 
The set of {\bf abstract formulas}, denoted ${\cal F}$, is defined inductively:
If $\alpha\in{\cal P}\cup\{\top,\bot\}$, then $\alpha\in{\cal F}$;
If $\alpha,\beta\in{\cal F}$, then $\alpha\wedge\beta\in{\cal F}$;
And if $\alpha\in{\cal F}$, then $\neg\alpha\in{\cal F}$.
\end{definition}

If $|{\cal A}| = |{\cal P}|$, then for each $\alpha \in {\cal L}$, there is a $\beta\in {\cal F}$ (and vice versa) such that $\alpha$ and $\beta$ are isomorphic (i.e. they have the same syntax tree except for the atom associated with the leaves). For example, $\tt \neg(x_1 \wedge x_2 \wedge x_3)$ is an abstract formula that is isomorphic to the AMR formula in Example \ref{ex:amrformula}.

If $\phi$ is an AMR formula, or an abstract formula, then ${\myatoms}(\phi)$ is the set of atoms used in $\phi$.
If $\Phi$ is a set of AMR formula, or a set of abstract formulas, then ${\myatoms}(\Phi)$ is the set of atoms used in the formulas in $\Phi$.

In order to consider how we can translate AMR formulas into abstract formulas, we need to consider when two atoms can be regarded as equivalent, and therefore mapped to the same propositional letter. In this section, we introduce the notion of a matching relation $\simeq$ between pairs of AMR formulas, and consider some options for defining this relation based on the syntax of the atoms in the AMR language, and then consider alternative definitions for this matching relation in the next subsection based on embeddings. 

We could define $\simeq$ as the standard equality $=$ relation. So for any $\alpha,\beta \in {\cal A}$, $\alpha\simeq\beta$ iff $\alpha = \beta$. However, because of the way that the AMR parser works, we want to make some further pairs of atom match. So we will use the following definition as our first option for the $\simeq$ relation.

\begin{definition}
\label{def:simpleequality}
For AMR formulas $\phi$ and $\psi$, 
a {\bf syntactic matching relation}, denoted $\simeq^{\phi,\psi}_{\sf sm}$, 
is defined by the following:
If $p(a) \in \myatoms(\phi)$ and $p(b)  \in \myatoms(\psi)$, 
then $p(a) \simeq^{\phi,\psi}_{\sf sm} p(b)$; 
And if $p(a), q(c), r(a,c) \in \myatoms(\phi)$
and $p(b), q(d), r(b,d)  \in \myatoms(\psi)$, 
then $r(a,c) \simeq^{\phi,\psi}_{\sf sm} r(b,d)$.
\end{definition}

So a pair of monadic atoms can be considered the same if they have the same predicate name without regard to the constant symbols, and two dyadic predicates can be considered the same if each constant symbol corresponds to the same monadic predicate relations. As we illustrate in the following example, $\tt car(c)$ occurring in the premise and $\tt car(c1)$ occurring in the claim can be considered the same because we are making the assumption the $\tt c$ and $\tt c1$ refer to the same car. In contrast, $\tt blue(c)$ in the premise and $\tt black(b4)$ occurring in the claim cannot be considered the same because $\tt b$ refers to blue and $\tt b4$ refers to black.

\begin{example}
Let $\phi$ = ${\tt car(c) \wedge ARG1(c,b) \wedge blue(b)}$, 
$\psi_1$ = ${\tt car(c1) \wedge ARG1(c1,b3) \wedge blue(b3)}$,
and $\psi_2$ = ${\tt car(c1)}$ $\wedge$ ${\tt ARG1(c1,b4)}$ $\wedge$ ${\tt black(b4)}$.
Therefore, for premise $\phi$ and claim $\psi_1$, 
${\tt car(c) \simeq^{\phi,\psi_1}_{\sf sm} car(c1)}$,
${\tt blue(b) \simeq^{\phi,\psi_1}_{\sf sm} blue(b3)}$
and ${\tt ARG1(c, b) \simeq^{\phi,\psi_1}_{\sf sm} ARG1(c1,b3)}$ are all holding.
In contrast, for premise $\phi$ and claim $\psi_2$, 
${\tt car(c) \simeq^{\phi,\psi_2}_{\sf sm} car(c1)}$,  
but ${\tt blue(b) \not\simeq^{\phi,\psi_2}_{\sf sm} black(b4)}$, 
and ${\tt ARG1(c, b) \not\simeq^{\phi,\psi_2}_{\sf sm} ARG1(c1,b4)}$.
\end{example}

\begin{definition}
\label{def:translation}
Given a set of AMR formulas $\Phi$, and an matching relation $\simeq$, 
the function $g: {\cal A} \rightarrow {\cal P}$ is a {\bf translation} for $\Phi$ and $\simeq$
iff for all $\phi,\psi\in\Phi$, 
for all $\alpha\in {\myatoms}(\phi)$, 
for all $\beta\in {\myatoms}(\psi)$,
$\alpha \simeq \beta \mbox{ iff } g(\alpha) = g(\beta)$.
Also for the tautology symbol, $g(\top) = \top$.
\end{definition}

The above definition ensures that if matching atoms are used in different AMR formulas in $\Phi$, 
then they are translated to the same atom in the abstract formulas.
If we were to use standard equality $=$ for $\simeq$, 
then a bijection from ${\cal A}$ to ${\cal P}$ would be an example of a translation function. 

\begin{example}
\label{ex:translation}
Consider $\phi_1 = {\tt car(c)} \wedge {\tt ARG1(c,r)} \wedge {\tt red(r)} \wedge {\tt ARG2(c,f)} \wedge {\tt fast(f)}$
and $\phi_2 = {\tt car(c1)} \wedge {\tt ARG1(c1,r1)} \wedge {\tt red(r1)}$.
Let $\Phi = \{\phi_1,\phi_2\}$.
A translation $g$ for $\Phi$ is 
$g({\tt car(c)}) = g({\tt car(c1)}) = {\tt x_1}$,
$g({\tt ARG1(c,r)}) = g({\tt ARG1(c1,r1)}) = {\tt x_2}$,
$g({\tt red(r)}) = g({\tt red(r1)}) = {\tt x_3}$,
$g({\tt ARG1(c,f)}) = {\tt x_4}$,
and $g({\tt fast(f)}) = {\tt x_5}$.
\end{example}

An AMR formula can be translated into an abstract formula by treating the abstract formulas as an abstraction.

\begin{definition}
\label{def:abstraction}
Let $\Phi$ be a set of AMR formula and let $g$ be a translation for $\Phi$.
For $\phi \in \Phi$, an {\bf abstraction} of $\phi$ is
$\myabstract_g(\phi)$ where $\myabstract_g$ is defined as follows:
(1) $\myabstract_g(\alpha\wedge\beta) = \myabstract_g(\alpha) \wedge \myabstract_g(\beta)$;
(2) $\myabstract_g(\neg\alpha) = \neg\myabstract_g(\alpha)$;
And (3) $\myabstract_g(\alpha) = g(\alpha)$ when $\alpha\in{\cal A}$.
For a set $\Phi$, $\myabstract_g(\Phi)$ = $\{\myabstract_g(\phi) \mid \phi \in \Phi\}$.
\end{definition}

\begin{example}
Continuing Example \ref{ex:translation},
an abstraction of $\phi_1$ is ${\tt x_1} \wedge {\tt x_2} \wedge {\tt x_3} \wedge {\tt x_4} \wedge {\tt x_5}$
and $\phi_2$ is ${\tt x_1} \wedge {\tt x_2} \wedge {\tt x_3}$.
\end{example}

\begin{example}
\label{e11}
For the premise sentence ``A man in an orange vest leans over a pickup truck", 
we have the premise 
$\tt \phi = \tt man(m) \wedge vest(v) \wedge orange(o)$
$\tt \wedge location(m,v) \wedge \tt mod(o,v) $
$\tt \wedge  lean\_over(l) \wedge ARG0(m,l) \wedge truck(t) \wedge ARG1(l,t)$, 
and for the claim sentence is ``A man is touching a truck.", we have the claim 
$\psi =  \tt  \tt man(m1) \wedge touch(t2) \wedge ARG0(m1,t2) \wedge truck(t1) \wedge ARG1(t2,t1)$. 

Let the translation function $g$ be defined as $g(\tt man(m))$ = $g(\tt man(m1))$ = $\tt x_1$, $g(\tt vest(v))$ = $\tt x_2$, $g(\tt orange(o))$ = $\tt x_3$,  $g(\tt location(m,v))$ = $\tt x_4$, $g(\tt mod(o,v))$ = $\tt x_5$, $g(\tt lean\_over(l))$ = $\tt x_6$, $g(\tt ARG0(m,l))$ = $\tt x_7$, $g(\tt truck(t))$ = $g(\tt truck(t1))$ = $\tt x_8$, $g(\tt ARG1(l,t))$ = $\tt x_9$,  $g(\tt touch(t2))$ = $\tt x_{10}$, $g(\tt ARG0(m1,t2))$ = $\tt x_{11}$, $g(\tt ARG1(t2,t1))$ = $\tt x_{12}$. 

So $\myabstract_g(\phi) = \{ \tt x_1 \wedge \tt x_2 \wedge \tt x_3 \wedge \tt x_4 \wedge \tt x_5\wedge \tt x_6\wedge \tt x_7\wedge \tt x_8 \wedge x_9\}$ and $\myabstract_g(\psi) = \tt x_1 \wedge x_{8} \wedge \tt x_{10} \wedge \tt x_{11} \wedge x_{12}$.

\end{example}

In the above example, the premises do not imply the claim. 
Yet common sense suggests that this entailment should hold. 
To address this, we introduce relaxed propositional entailment next.

\subsubsection{Relaxed Propositional Entailment}

In this section, we define the $\simeq$ relation using sentence embeddings. Essentially, if for two AMR atoms $\alpha$ and $\beta$, 
the similarity between the items of text corresponding to the two atoms is greater than a threshold $\tau$, we treat them as equivalent.

An {\bf embedding} of a string (i.e. a substring of an input sentence) $S$ is a vector $v$ given by an injective mapping function $f$ (i.e. $f(S) = v$). 
In our pipeline, the $f$ function is the sentence transformer paraphrase-multilingual-MiniLM-L12-v2  \cite{reimers-2019-sentence-bert} that encodes a word or a substring as high dimension vector.

We use the commonly-used cosine similarity definition for measuring  {\bf similarity} between two embedding vectors $v_1$ and $v_2$ as follows.
\[
\similarity(v_1,v_2) = cos(\theta) = \frac{v_1 \cdot v_2}{||v_1||||v_2||}
\]
where $\theta$ is the angle between the vectors, $v_1 \cdot v_2$ is dot product between $v_1,v_2$, and $||v_1||$ represents the L2 norm.

To apply the embedding, we first extract a substring from the input sentence based on the dyadic atoms in the AMR formula, and combining two argument entries directly if no substring is found, as defined next. 

\begin{definition}
Let $\phi$ be the AMR formula that represents sentence $S$,
and we treat $S$ as a sequence of $n$ words $[ w_1,\ldots,w_n ]$. 
The {\bf substring} function, denoted $\mysubstring_S$, is defined as follows
for $r(a,b)$ when $p(a),q(b),r(a,b) \in {\myatoms}(\phi)$: 
If there are words $w_i$ and $w_j$ in $S$ 
where the stem of $p$ equals the stem of $w_i$, 
and the stem of $q$ equals the stem of $w_j$, 
then $\mysubstring_S(r(a,b)) = [ w_i,\ldots,w_j ]$, 
otherwise, $\mysubstring_S(r(a,b)) = [ p,q ]$.
\end{definition}

After we extract the corresponding substrings, we get their sentence embeddings, as illustrated next.

\begin{example}
\label{e13}
Continuing Example \ref{e11}, the premise sentence is $S$ = [A, man, in, an, orange, vest, leans, over, a, pickup, truck], and so $\mysubstring_S(\tt ARG0(m,l))$  = [man, lean\_over] and $\mysubstring_S(\tt ARG1(l,t))$  = [lean\_over,truck], 
as the word ``lean\_over" is not in $S$. 
The claim sentence is $S'$ = [A, man, is, touching, a, truck], 
and so $\mysubstring_{S'}(\tt ARG0(m1,t2))$  = [man, is, touching] 
and $\mysubstring_{S'}(\tt ARG1(t2,t1))$  = [touching,a,truck], 
since touching and touch have the same stem.

\end{example}

\begin{definition}
For an AMR formula $\phi$ that represents sentence $S$, 
an {\bf extraction function}, denoted $\extract_S$, 
is a function 
such that if $p(a)\in {\myatoms}(\phi)$,
then $\extract_S(p(a)) = p$,
and if $r(a,b)\in {\myatoms}(\phi)$,
then $\extract_S(r(a,b)) = \mysubstring_S(r(a,b))$.
\end{definition}

\begin{definition}
For premise $\phi$ (resp. claim $\psi$) 
that represents sentence $S$ (resp $S'$),
the {\bf neuro-matching relation}, 
denoted $\simeq^{S,S'}_n$, is defined as follows, 
where $\tau\in [0,1]$ is the {\bf neuro-matching threshold}, 
and $\beta\in {\myatoms}(\psi)$.
\[
\begin{array}{l}
\alpha \simeq^{S,S'}_n \beta \mbox{ iff }  (\alpha,x) \in {\similaratoms}(\beta)\\
\hspace{15mm} \mbox{ and for all } (\alpha',y) \in {\similaratoms}(\beta), x \geq y
\end{array}
\]
where ${\similaratoms}(\beta)$ = $\{ (\alpha,x) \mid \alpha\in {\myatoms}(\phi)$ 
and $x > \tau$
and ${\similarity}({ f}(\extract_{S}(\alpha)),{ f}(\extract_{S'}(\beta))) = x\}$ 
\end{definition}

This definition finds the best match in the premise (if there is one with a similarity greater than the relaxation threshold) for each atom in the claim: So $\alpha \simeq^{S,S'}_n \beta$ holds when for all premise atoms $\alpha_1,\ldots,\alpha_n$, with similarity scores with $\beta$ of $s_1,\ldots,s_n$ respectively, 
then $\alpha$ is the atom $\alpha_i$ for which $s_i = \max(s_1,\ldots,s_n)$. 

In relaxed propositional entailment, we use the following combined-matching relation in Definition \ref{def:translation}. 

\begin{definition}
\label{ref:combinedmatch}
Let $\phi$ (resp. $\psi$) be the AMR formula that represents premise $S$ (resp claim $S'$),
and let $\alpha\in {\myatoms}(\psi)$, $\beta \in {\myatoms}(\phi)$. 
The {\bf combined-matching relation}, 
denoted $\simeq^{S,S'}_{\sf cm}$, is defined as $\alpha \simeq^{S,S'}_{\sf cm} \beta$ 
iff $\alpha \simeq^{\phi,\psi}_{\sf sm} \beta$ or $\alpha \simeq^{S,S'}_{\sf nm} \beta$. 
\end{definition}

\begin{example}
If we have an atom $\tt automobile(a)$ (resp. $\tt automobile(a1)$) in the AMR formula for the premise $S$ (resp. claim $S'$)
then $\tt automobile(a) \simeq^{S,S'}_{\sf sm} automobile(a1)$ holds.
Whereas if $\tt car(c)$ is an atom in the claim, then we consider the neuro-matching relation.
Suppose ${\similarity}(f(\extract_S({\tt automobile(a)})),f(\extract_{S'}({\tt car(c)}))) 
={\similarity}(f({\tt automobile}),f({\tt car})) = 0.68$, and $\tau = 0.6$, 
then $\tt automobile(a) \simeq^{S,S'}_{\sf nm} car(c)$ holds. 
In both cases, the combined-matching relation holds.
\end{example}

As well as using combined matching in the definition for translation (i.e. Definition \ref{def:translation}), we can simplify the claim based on the premises using the following definition. Essentially, if we have the neuro-matching of a dyadic predicate, then we can say the corresponding monadic predicates are satisfiable, which is done by making them true (represented by the $\top$ tautology symbol).

\begin{definition}
\label{def:simplification}
Let $\phi$ be a premise and $\psi$ be a claim, 
where $p(a), q(b), r(a,b)\in {\myatoms}(\psi)$. 
If there exists $\alpha \in {\myatoms}(\phi)$ 
s.t. $\alpha \simeq^{S,S'}_{\sf nm} r(a,b)$,
then the {\bf claim simplification} is to let 
$p(a) = \top$
and $q(b) = \top$.
\end{definition}

\begin{example}
\label{e15}
Continuing Example \ref{e13}, suppose $\tau = 0.6$, and
\[
\begin{array}{l}
{\similarity}(f(\extract_S({\tt ARG0(m,l)}) ,f(\extract_{S'}({\tt ARG0(m1,t2)})))) =\\
{\similarity}(f({\tt man\ leans\_over}) ,f({\tt man\ is\ touching})) = 0.7
\end{array}
\]
then $\tt ARG0(m,l) \simeq^{S,S'}_{\sf nm} \tt ARG0(m1,t2)$
and $\tt ARG0(m,l) \simeq^{S,S'}_{\sf cm} \tt ARG0(m1,t2)$.
So $g({\tt ARG0(m,l)}) = g(\tt ARG0(m1,t2)) = \tt x_7$. 
And from Definition \ref{def:simplification}, 
$g(\tt man(m1))= \top$ and $g(\tt touch(t2))= \top$.
Also suppose
\[
\begin{array}{l}
{\similarity}(f(\extract_S({\tt ARG1(l,t)}) ,f(\extract_{S'}({\tt ARG1(t2,t1)}))) =\\
{\similarity}(f({\tt leans\_over\ truck}),f({\tt touching\ a\ truck})) > \tau
\end{array}
\]
So we have $g({\tt ARG1(l,t)}) = g({\tt ARG1(t2,t1)}) = \tt x_9$, $g(\tt touch(t2))=\top$, and $g(\tt truck(t1))=\top$ holding.
\end{example}

Given a premise $\phi$ and a claim $\psi$,
let $\phi'$ (resp. $\psi'$) be the
abstraction (Definition \ref{def:abstraction}) of $\phi$ (resp. $\psi$) 
obtained from applying
the translation function (Definition \ref{def:translation}) 
with the combined-matching relation (Definition \ref{ref:combinedmatch}) 
and then applying claim simplification (Definition \ref{def:simplification}).
We call  $\phi'$ (resp. $\psi'$) the {\bf relaxed premise} (resp. {\bf relaxed claim}).
When  $\phi'$ entails $\psi'$ 
using propositional entailment (i.e. $\{\phi'\}\vdash\psi'$ holds), 
then we say that $\phi$ entails $\psi$ using {\bf relaxed propositional entailment}. 

\begin{example}
Returning to Example \ref{e11}, 
we use Example \ref{e15}, 
so that the relaxed premise $\phi'$ is 
$\tt x_1 \wedge \tt x_2 \wedge \tt x_3 \wedge \tt x_4 \wedge \tt x_5\wedge \tt x_6\wedge \tt x_7\wedge \tt x_8 \wedge \tt x_9$.
and the relaxed claim is 
$\psi'$ = $\top \wedge \top \wedge \top \wedge x_7 \wedge x_9$. 
Hence $\phi'$ entails $\psi'$, 
and so $\phi$ entails $\psi$ using relaxed propositional entailment.
\end{example}

Since $\phi$ entails $\psi$ using propositional entailment, $\phi$ entails $\psi$ using propositional entailment. So, relaxed propositional entailment is more permissive. 

\subsection{Contradiction methods}
\label{section:contradiction}

We use the relaxed propositional entailment for identifying whether a premise contradicts a claim. We also require a notion of forgetting explained next.

Often explanations express things that cannot happen. This appears in the logical representation as the negation of a conjunction of atoms. For example, the explanation ``you cannot walk when you sleep on the bed" might be represented by $\tt \neg(possible(p) \wedge ARG1(p, w) \wedge walk(w) \wedge ARG0(w, y) \wedge you(y) \wedge time(p, s) \wedge sleep(s) \wedge ARG0(s, y) \wedge location(s, b) \wedge bed(b))$. 
Suppose we have the premise ``you are sleeping" which is represented by $\tt sleep(s1) \wedge ARG0(s1, y1) \wedge you(y1)$. 
When we try to prove that the premise contradicts the claim ``you are walking" which is represented 
by $\tt walk(w2) \wedge ARG0(w2, y2) \wedge you(y2)$ with the help of the explanation, 
the additional atoms $\tt possible(p)$, $\tt time(p, s)$, $\tt ARG1(p, w)$, $\tt location(s, b)$, $\tt bed(b)$ are all irrelevant. 
This irrelevancy means we are unable to show the conjunction of premise and explanation are inconsistent with the claim. 
To address this, we introduce a method to forget irrelevant atoms.

\begin{example}
Consider the premise $\neg (x_1 \wedge x_2 \wedge x_3)$  and claim $\tt x_1 \wedge x_2$ as abstract formulas. The conjunction $\tt x_1 \wedge x_2 \wedge \neg (x_1 \wedge x_2 \wedge x_3)$ is consistent as there is an irrelevant atom $\tt x_3$. 
If we forget the atom $\tt x_3$ as it is not in the claim, 
we have $\tt x_1 \wedge x_2 \wedge \neg (x_1 \wedge x_2)$ which is inconsistent.
\end{example}

The forgetting method ignores the atoms in the premises that are not in the claim as defined next.

\begin{definition}
For premise $\phi'$ and claim $\psi'$,
the {\bf forgettable atoms} of $\phi'$ and $\psi'$ 
are $\myatoms(\phi')\setminus\myatoms(\psi')$. 
\end{definition}

\begin{example}
Suppose $\myatoms(\phi') = \{\tt x_1,x_2,x_3,x_4\}$ 
and  $\myatoms(\psi') = \{\tt x_1,x_2\}$. 
The forgettable atoms of $\phi'$ and $\psi'$ 
are $\tt \{x_1,x_2,x_3,x_4\}\setminus\{x_1,x_2\} = \{x_3,x_4\}$.
\end{example}

 \begin{definition}
For a forgettable atom $\alpha'$, the {\bf forget function}, $\myforget_{\alpha'}$, is defined as follows, 
where $\phi'$ and $\psi'$ are abstract formulas:
(1) $\myforget_{\alpha'}(\phi' \wedge \psi') = \myforget_{\alpha'}(\phi') \wedge \myforget_{\alpha'}(\psi')$;
(2) $\myforget_{\alpha'}(\neg \beta') = \neg \myforget_{\alpha'}(\beta')$;
(3) $\myforget_{\alpha'}(\alpha') = \top$;
and
(4) $\myforget_{\alpha'}(\beta') = \beta'$ if $\beta'$ is an atom and $\beta' \neq \alpha'$.
The forget function is generalized to a set of forgettable atoms $\Gamma = \{ \alpha'_1, \ldots, \alpha'_n \}$ as follows: 
$\myforget_{\Gamma}
= 
\myforget_{\alpha'_1}(......\myforget_{\alpha'_n}(\phi')..)$. 
 \end{definition}

\begin{example}
\label{el}
    Suppose $\tt \alpha' = x_1, \phi' = x_1 \wedge x_2 \wedge x_3, \psi' = x_4 \wedge x_5, \beta = x_6$, 
    $\myforget_{\alpha'}(\phi' \wedge \psi') = \myforget_{\alpha'}(\phi') \wedge \myforget_{\alpha'}(\psi') 
    = \myforget_{\tt x_1}(\tt x_1 \wedge x_2 \wedge x_3) \wedge \myforget_{x_1}(x_4 \wedge x_5) 
    = x_2 \wedge x_3 \wedge x_4 \wedge x_5$; $\myforget_{\alpha'}(\neg \beta') = \neg \myforget_{\alpha'}(\beta') = \neg \myforget_{\tt x_1}({\tt x_6}) = \tt \neg x_6$
\end{example}

The forgetting method allows more pairs of premise and claim to be classed as inconsistent. 
In our empirical study, we investigate the trade-off of recall and precision. 

\subsection{Classification of relations}
\label{section:classification}

The aim of our pipeline is to identify the relationship between a premise $\phi$ and a claim $\psi$. We use automated reasoning to check both entailment and contradiction simultaneously, and then classify according to Table \ref{table:classification}.

\begin{table}[t]
\centering
\begin{tabular}{ | m{6em} | m{1.5cm}| m{1.5cm}|  } 
  \hline
   Predicted class & $\phi' \land \neg \psi'$   & $\phi^* \land \psi'$ \\ 
  \hline
 Entailment & False & True  \\ 
  \hline
  Contradiction & True & False \\ 
  \hline
  Neutral & True & True \\ 
  \hline
\end{tabular}
\caption{\label{table:classification}Classification of relations. 
Let $\phi$ (resp. $\psi$) be a premise (resp. claim),
let $\phi'$ (resp. $\psi'$) be the relaxed premise (resp. relaxed claim),
and let $\phi^{\ast}$ be $\myforget_{\Gamma}(\phi')$ where $\Gamma$ is the set of forgettable atoms of $\phi'$ and $\psi'$. 
For columns 2 and 3, PySAT returns true if the formula is consistent, and returns False if it is inconsistent. 
So for instance the entailment relation holds
when $\phi' \land \neg \psi'$ is False  and $\phi^* \land \psi'$ is True.
}
\end{table}

There is another possibility for classification of relations, and that is when $\phi' \wedge \neg \psi'$ is inconsistent (False) and $\phi^* \wedge \psi'$ is inconsistent (False). 
This would arise if $\phi$ is inconsistent. We investigated this by randomly selecting 1000 data items from the e-SNLI dataset, and in all cases, $\phi$ was consistent. 
This suggests that inconsistent premises $\phi$ are a rare or potentially non-existent occurrence in the datasets we used for our studies. 
We may also get $\phi' \wedge \neg \psi'$ and $\phi^* \wedge \psi'$ being False due to the forgetting method. 
Consider premise $\phi' = \tt x_1 \wedge \neg x_2$ and claim $\psi' = \tt x_1$. 
So $\phi' \wedge \neg \psi'$ is inconsistent (False). 
Because $\tt x_2$ is a forgettable atom,  
we have $\phi^* = \tt x_1 \wedge \neg x_2 = \tt x_1 \wedge \neg \top  = \tt x_1 \wedge \bot$, 
so $\phi^* \wedge \psi = \tt x_1 \wedge \bot \wedge x_1$ is also inconsistent (False). 
In the following experiments, 
we eliminate examples for which both $\phi \wedge \neg \psi$ and 
$\phi^* \wedge \psi$ are inconsistent (False), 
as a preprocessing step (as explained below), 
since they cannot be classified as any of the three classes (of entailment, contradiction, or neutral).

\section{Results}

We use the recall, precision, accuracy and F1-score as the evaluation metric. 
In the results, accuracy refers to the overall accuracy encompassing all three classes, and support refers to the number of examples in the class that we used in the experiment (which is 500 for each class in each experiment) {\em minus} the number of those examples that were false for both entailment and contradiction (i.e. both $\phi' \wedge \neg \psi'$ and 
$\phi^* \wedge \psi'$ are inconsistent (False)). In most of experiments, support was above 98\%. Support dropped to 80\% for the neutral class in the results reported in Table \ref{table:esnli:withforget}. 

\begin{table}[t]
\centering
\subfloat[Explanations and no forgetting (accuracy 0.553)\label{table:esnli:noforget}]
{
\begin{tabular}{ | m{5.5em} | m{1.2cm}| m{1cm} |m{1cm} |m{1cm} | } 
  \hline
Class & Precision   & Recall & F1 & Support\\ 
  \hline
 Entailment & 0.604   &    0.928  & 0.732 & 500\\ 
  \hline
  Contradiction &  0.538   &   0.014 & 0.027 & 498\\ 
  \hline
  Neutral & 0.499   &   0.716  & 0.588 & 497\\ 
  \hline
\end{tabular}
}

\centering
\subfloat[Explanations and forgetting (accuracy 0.560)\label{table:esnli:withforget}]
{
\begin{tabular}{ | m{5.5em} | m{1.2cm}| m{1cm} |m{1cm} | m{1cm} |} 
  \hline
Class & Precision   & Recall & F1 & Support \\ 
  \hline
 Entailment & 0.734   &    0.926  & 0.819& 488\\ 
  \hline
  Contradiction &  0.450   &   0.640 & 0.529 & 455 \\ 
  \hline
  Neutral & 0.111   &   0.023  & 0.037& 400 \\ 
  \hline
\end{tabular}
}

\centering
\subfloat[Forgetting and no explanation (accuracy 0.456)\label{table:esnli:withoutexplanation}]
{
\begin{tabular}{ | m{5.5em} | m{1.2cm}| m{1cm} |m{1cm} |m{1cm} | } 
  \hline
 Class & Precision   & Recall & F1 & Support \\ 
  \hline
 Entailment & 0.515   &    0.786  & 0.622 & 495\\ 
  \hline
  Contradiction &  0.652   &   0.030 & 0.057 & 499 \\ 
  \hline
  Neutral & 0.387   &   0.554  & 0.456 &500 \\ 
  \hline
\end{tabular}
}
\caption{\label{table:esnli}Results from the e-SNLI dataset}
\end{table}

\subsection{Different neuro-matching thresholds}
\label{section:similarity}

\begin{figure*}[t]
\centering
\subfloat[Entailment]{\includegraphics[scale=.35]{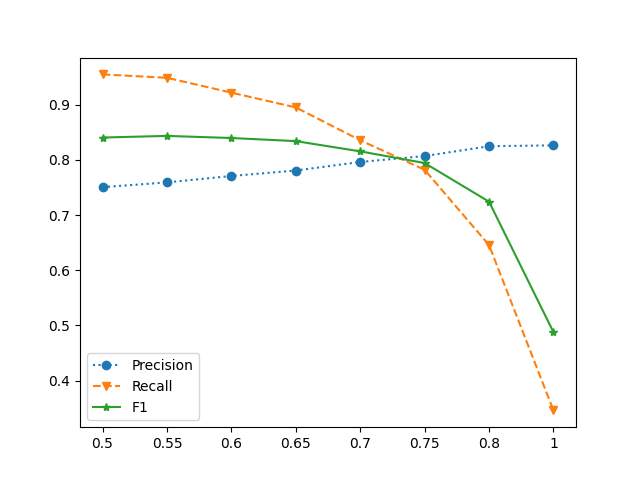}}
\subfloat[Contradiction]{\includegraphics[scale=.35]{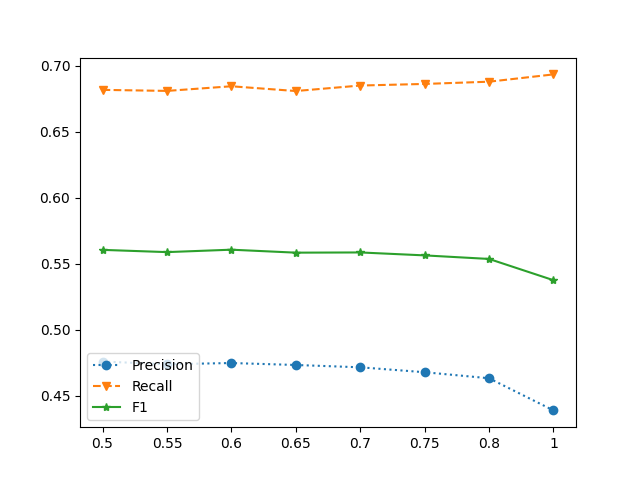}}
\subfloat[Neutral]{\includegraphics[scale=.35]{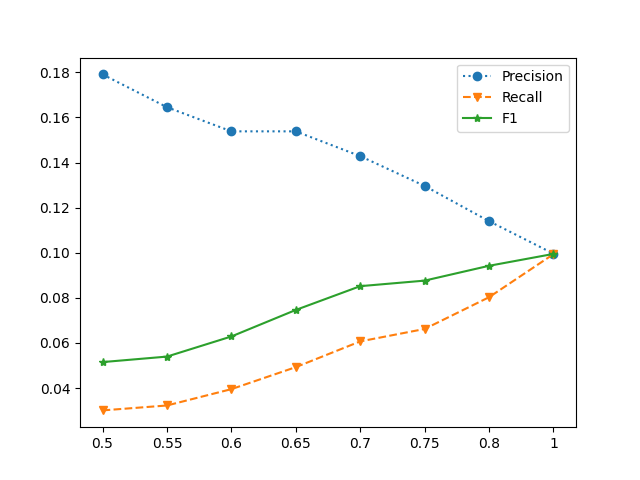}}
    \caption{Results concerning the neuro-matching relaxation threshold. The y-axis represents the scores of precision, recall, F1-score, and accuracy, and the x-axis represents the threshold $\tau$.}
\label{f5}
\end{figure*}

This section investigates how the performance changes with different neuro-matching thresholds $\tau$. We assume that the maximum sentence length is 20 words, we consider eight similarity thresholds $\tau$ (0.5-0.8 at 0.05 intervals, and 1); and we randomly select 1500 data items with three labels (500 each) from e-SNLI. 

In Figure \ref{f5}, we see that recall of entailment decreases gradually 
when $\tau$ increases from 0.55 to 0.7; 
and then decreases significantly when the $\tau$ increases above 0.7, 
whereas the precision of entailment rises slightly as $\tau$ increases.
The recall of contradiction is little affected by changes in $\tau \leq 0.8$,
whereas precision of contradiction gradually decreases as $\tau$ increases.
The recall of neutrality rises slightly as $\tau \leq 0.8$ increases,
whereas precision for neutrality drops markedly as $\tau  \leq 0.8$ increases. 

When $\tau$ = 1, the pipeline in effect has no relaxation method. The performance on the entailment class decreases dramatically compared to when relaxation methods are employed. This suggests that relaxation methods are effective at improving performance for identifying entailment and contradiction, but this comes at the price of substantial reduction in the precision in identifying the neutral class. 

\subsection{Pipeline with forgetting versus without forgetting}

This section investigates the performance of the pipeline with, or without, the forgetting method. We assume that there is no max length of each sentence, $\tau$ is 0.55 and we randomly select 1500 data items, with three labels (500 each), from the e-SNLI dataset.

In Tables \ref{table:esnli:noforget} and \ref{table:esnli:withforget}, we can see that the recall of the contradiction class increases significantly in the pipeline with the forgetting method compared to the pipeline without it. The recall for the neutral class decreased by 0.694, because the forgetting methods made it more challenging for the pipeline to correctly identify the neutral class, but the improvement in identifying the contradiction class, and the maintenance of overall accuracy, suggests that the use of the forgetting method is valuable in the pipeline.

\subsection{Performance on the three datasets}
\label{section:performance}

This section investigates the performance with three datasets described earlier. We assume no max length for each sentence, $\tau$ is 0.55, and we randomly select 1500 data items with label entailment, contradiction and neutral (500 each). 

Tables \ref{table:esnli:withforget} and \ref{table:esnli:withoutexplanation} 
give results for e-SNLI.
``No explanation'' means only the premise and claim are used to make the prediction,
whereas "Explanations" means conjoining each premise with its explanation. 
We can see that the performance improves with the use of the explanations, especially for the recall of entailment and contradiction classes. The precision of the entailment class also increases, which means the explanation helps the pipeline classify less data incorrectly as entailment. However, the forgetting method reduces recall of the neutral relation.

Our pipeline performs best with the SICK dataset, even though there are no explanations available (Table \ref{table:sick}). This is because each sentence in the SICK dataset contains more explicit information and requires less common sense knowledge. Though, recall of neutral is still not high. 

Finally, we can see that the pipeline performs worst with the MultiNLI dataset (Table \ref{table:mnli}).
The MultiNLI results are similar to the e-SNLI results without explanation. We believe that with the MultiNLI dataset, we would require extra common sense knowledge in order to be able to improve the performance in identifying the relations. 

\begin{table}[t]
\centering
\begin{tabular}{ | m{5.5em} | m{1.2cm}| m{1cm} |m{1cm} |m{1cm} | } 
  \hline
 Class & Precision   & Recall & F1 & Support\\ 
  \hline
 Entailment & 0.659   &    0.942  & 0.776 & 499\\ 
  \hline
  Contradiction &  0.906   &   0.810 & 0.855 & 500  \\ 
  \hline
  Neutral & 0.826  &   0.560  & 0.667 & 500 \\ 
  \hline
\end{tabular}
\caption{\label{table:sick}SICK dataset with forgetting (accuracy 0.771)}
\end{table}

\begin{table}[t]
\centering
\begin{tabular}{ | m{5.5em} | m{1.2cm}| m{1cm} |m{1cm} | m{1cm} | } 
  \hline
Class & Precision   & Recall & F1 & Support\\ 
  \hline
 Entailment & 0.610   &    0.578  & 0.593 & 476\\ 
  \hline
  Contradiction &  0.552   &   0.374 & 0.446 &494\\ 
  \hline
  Neutral & 0.459  &   0.629  & 0.531 & 496\\ 
  \hline
\end{tabular}
\caption{\label{table:mnli}MultiNLI dataset with forgetting (accuracy 0.527)}
\end{table}

For comparison, 
a DL-based seq2seq attention model with e-SNLI data has accuracy 0.8171 \cite{NEURIPS2018_4c7a167b},
a BERT-based model with SICK data has accuracy 0.8674 \cite{hu-etal-2020-monalog},
and a DL-based BiLSTM model with MultiNLI has accuracy 0.669 \cite{N18-1101}. 
So our pipeline does not perform as well. 
However, a key advantage of our approach is that it does not need training, and it produces output in the form of logical formulas. Both the original formulas, and relaxed formulas can be given as part of an explanation. 
Our pipeline has similar results with MultiNLI as other logic-inference-based methods, but they have the disadvantage of needing extra information to deal with sentences involving common sense knowledge.  

\section{Discussion}

In this paper, we propose a pipeline that combines a pre-trained AMR parser with automated reasoning based on a SAT solver to provide an explainable approach to identifying entailment and contradiction relationships between a premise sentence and a claim sentence. From the evaluation with the three RTE datasets, we obtain a good trade-off of precision and recall. We believe that our neurosymbolic approach is useful for analysing arguments, where it is often infeasible to use normal symbolic reasoning because of the need to handle common sense reasoning. In future work, we will use our pipeline for premises and claims obtained by argument mining \cite{Lawrence2019} in order to generate logical arguments, check their validity and consistency, and check support (resp. attack) relationships specified in terms of entailment (resp. contradiction). Since most natural language arguments are enthymemes, i.e. they involve implicit knowledge (common and common sense), we also want to investigate how existing proposals for logic-based understanding of enthymemes, such as \cite{Hunter2022}, could be combined with our approach. 

\bibliography{aaai25}

\begin{thebibliography}{31}
\providecommand{\natexlab}[1]{#1}

\bibitem[{Alabbas(2013)}]{phdthesis}
Alabbas, M. 2013.
\newblock \emph{Textual Entailment for Modern Standard Arabic}.
\newblock Ph.D. thesis, University of Manchester.

\bibitem[{Atkinson et~al.(2017)Atkinson, Baroni, Giacomin, Hunter, Prakken,
  Reed, Simari, Thimm, and Villata}]{Atkinson2017}
Atkinson, K.; Baroni, P.; Giacomin, M.; Hunter, A.; Prakken, H.; Reed, C.;
  Simari, G.~R.; Thimm, M.; and Villata, S. 2017.
\newblock Towards Artificial Argumentation.
\newblock \emph{{AI} Magazine}, 38(3): 25--36.

\bibitem[{Banarescu et~al.(2013)Banarescu, Bonial, Cai, Georgescu, Griffitt,
  Hermjakob, Knight, Koehn, Palmer, and
  Schneider}]{banarescu-etal-2013-abstract}
Banarescu, L.; Bonial, C.; Cai, S.; Georgescu, M.; Griffitt, K.; Hermjakob, U.;
  Knight, K.; Koehn, P.; Palmer, M.; and Schneider, N. 2013.
\newblock {A}bstract {M}eaning {R}epresentation for Sembanking.
\newblock In \emph{Proc. of Linguistic Annotation Workshop and Interoperability
  with Discourse}, 178--186. ACL.

\bibitem[{{Ben-Sghaier}, Bakari, and Neji(2020)}]{10.1007/978-3-030-16660-1_40}
{Ben-Sghaier}, M.; Bakari, W.; and Neji, M. 2020.
\newblock Arabic Logic Textual Entailment with Feature Extraction and
  Combination.
\newblock In \emph{Intelligent Systems Design and Applications}, 400--409.
  Springer.

\bibitem[{Bos(2016)}]{bos-2016-squib}
Bos, J. 2016.
\newblock {S}quib: Expressive Power of {A}bstract {M}eaning {R}epresentations.
\newblock \emph{Computational Linguistics}, 42(3): 527--535.

\bibitem[{Bos and Markert(2005)}]{10.3115/1220575.1220654}
Bos, J.; and Markert, K. 2005.
\newblock Recognising Textual Entailment with Logical Inference.
\newblock In \emph{Proc. of HLT}, 628–635. ACL.

\bibitem[{Bowman et~al.(2015)Bowman, Angeli, Potts, and
  Manning}]{DBLP:journals/corr/BowmanAPM15}
Bowman, S.~R.; Angeli, G.; Potts, C.; and Manning, C.~D. 2015.
\newblock A Large Annotated Corpus for Learning Natural Language Inference.
\newblock \emph{CoRR}, abs/1508.05326.

\bibitem[{Camburu et~al.(2018)Camburu, Rockt\"{a}schel, Lukasiewicz, and
  Blunsom}]{NEURIPS2018_4c7a167b}
Camburu, O.-M.; Rockt\"{a}schel, T.; Lukasiewicz, T.; and Blunsom, P. 2018.
\newblock e-SNLI: Natural Language Inference with Natural Language
  Explanations.
\newblock In \emph{Advances in Neural Information Processing Systems},
  volume~31. Curran Assoc.

\bibitem[{Chanin and Hunter(2023)}]{chanin2023neuro}
Chanin, D.; and Hunter, A. 2023.
\newblock Neuro-symbolic Commonsense Social Reasoning.
\newblock \emph{arXiv preprint arXiv:2303.08264}.

\bibitem[{Dagan, Glickman, and Magnini(2006)}]{dagan2006machine}
Dagan, I.; Glickman, O.; and Magnini, B. 2006.
\newblock The PASCAL Recognising Textual Entailment Challenge.
\newblock In \emph{Machine Learning Challenges. Evaluating Predictive
  Uncertainty, Visual Object Classification, and Recognising Tectual
  Entailment}. Springer.

\bibitem[{Glickman, Dagan, and Koppel(2005)}]{10.1007/11736790_16}
Glickman, O.; Dagan, I.; and Koppel, M. 2005.
\newblock A Lexical Alignment Model for Probabilistic Textual Entailment.
\newblock In \emph{Proc. of Machine Learning Challenges Workshop}, 287–298.
  Springer-Verlag.

\bibitem[{Hovy et~al.(2006)Hovy, Marcus, Palmer, Ramshaw, and
  Weischedel}]{hovy-etal-2006-ontonotes}
Hovy, E.; Marcus, M.; Palmer, M.; Ramshaw, L.; and Weischedel, R. 2006.
\newblock {O}nto{N}otes: The 90{\%} Solution.
\newblock In \emph{Proc. of HLT}, 57--60. ACL.

\bibitem[{Hu et~al.(2020)Hu, Chen, Richardson, Mukherjee, Moss, and
  Kuebler}]{hu-etal-2020-monalog}
Hu, H.; Chen, Q.; Richardson, K.; Mukherjee, A.; Moss, L.~S.; and Kuebler, S.
  2020.
\newblock {M}ona{L}og: a Lightweight System for Natural Language Inference
  Based on Monotonicity.
\newblock In \emph{Proc. of Soc. for Comp. in Ling.}, 334--344. ACL.

\bibitem[{Hunter(2022)}]{Hunter2022}
Hunter, A. 2022.
\newblock Understanding Enthymemes in Deductive Argumentation Using Semantic
  Distance Measures.
\newblock In \emph{Proceedings of {AAAI}}, 5729--5736. {AAAI} Press.

\bibitem[{Ignatiev, Morgado, and Marques{-}Silva(2018)}]{imms-sat18}
Ignatiev, A.; Morgado, A.; and Marques{-}Silva, J. 2018.
\newblock {PySAT:} {A} {Python} Toolkit for Prototyping with {SAT} Oracles.
\newblock In \emph{SAT}, 428--437.

\bibitem[{Jijkoun, de~Rijke et~al.(2005)}]{jijkoun2005recognizing}
Jijkoun, V.; de~Rijke, M.; et~al. 2005.
\newblock Recognizing Textual Entailment using Lexical Similarity.
\newblock In \emph{Proc. of PASCAL Challenges Workshop on Recognising Textual
  Entailment}, 73--76.

\bibitem[{Kasper(1989)}]{kasper-1989-flexible}
Kasper, R.~T. 1989.
\newblock A Flexible Interface for Linking Applications to {P}enman{'}s
  Sentence Generator.
\newblock In \emph{Proc. of Workshop on Speech and Natural Language}.

\bibitem[{Kingsbury and Palmer(2002)}]{kingsbury-palmer-2002-treebank}
Kingsbury, P.; and Palmer, M. 2002.
\newblock From {T}ree{B}ank to {P}rop{B}ank.
\newblock In \emph{Proc. of LREC}. European Language Resources Association.

\bibitem[{Langkilde and
  Knight(1998)}]{langkilde-knight-1998-generation-exploits}
Langkilde, I.; and Knight, K. 1998.
\newblock Generation that Exploits Corpus-Based Statistical Knowledge.
\newblock In \emph{Proc. of ACL}, 704--710. ACL.

\bibitem[{Lawrence and Reed(2019)}]{Lawrence2019}
Lawrence, J.; and Reed, C. 2019.
\newblock Argument Mining: A Survey.
\newblock \emph{Computational Linguistics}, 45(4): 765--818.

\bibitem[{Lee et~al.(2021)Lee, Astudillo, Hoang, Naseem, Florian, and
  Roukos}]{DBLP:journals/corr/abs-2112-07790}
Lee, Y.; Astudillo, R.~F.; Hoang, T.~L.; Naseem, T.; Florian, R.; and Roukos,
  S. 2021.
\newblock Maximum Bayes Smatch Ensemble Distillation for {AMR} Parsing.
\newblock \emph{CoRR}, abs/2112.07790.

\bibitem[{Marelli et~al.(2014)Marelli, Bentivogli, Baroni, Bernardi, Menini,
  and Zamparelli}]{marelli-etal-2014-semeval}
Marelli, M.; Bentivogli, L.; Baroni, M.; Bernardi, R.; Menini, S.; and
  Zamparelli, R. 2014.
\newblock {S}em{E}val-2014 Task 1.
\newblock In \emph{Proc. of International Workshop on Semantic Evaluation},
  1--8. ACL.

\bibitem[{Meurer and {\em et. al.}(2017)}]{10.7717/peerj-cs.103}
Meurer, A.; and {\em et. al.} 2017.
\newblock SymPy: Symbolic Computing in Python.
\newblock \emph{PeerJ Computer Science}, 3: e103.

\bibitem[{Olausson et~al.(2023)Olausson, Gu, Lipkin, Zhang, Solar-Lezama,
  Tenenbaum, and Levy}]{olausson-etal-2023-linc}
Olausson, T.; Gu, A.; Lipkin, B.; Zhang, C.; Solar-Lezama, A.; Tenenbaum, J.;
  and Levy, R. 2023.
\newblock {LINC}: A Neurosymbolic Approach for Logical Reasoning by Combining
  Language Models with First-Order Logic Provers.
\newblock In \emph{Proc. of EMNLP}, 5153--5176. ACL.

\bibitem[{Padó et~al.(2015)Padó, Noh, Stern, Wang, and Zanoli}]{Padó2015167}
Padó, S.; Noh, T.-G.; Stern, A.; Wang, R.; and Zanoli, R. 2015.
\newblock Design and Realization of a Modular Architecture for Textual
  Entailment.
\newblock \emph{Nat. Lang. Eng.}, 21(2): 167 – 200.

\bibitem[{Pan et~al.(2023)Pan, Albalak, Wang, and Wang}]{pan-etal-2023-logic}
Pan, L.; Albalak, A.; Wang, X.; and Wang, W. 2023.
\newblock Logic-{LM}: Empowering Large Language Models with Symbolic Solvers
  for Faithful Logical Reasoning.
\newblock In \emph{Findings of EMNLP}, 3806--3824. ACL.

\bibitem[{Putra, Siahaan, and Saikhu(2024)}]{PUTRA2023}
Putra, I. M.~S.; Siahaan, D.; and Saikhu, A. 2024.
\newblock Recognizing Textual Entailment: A Review of Resources, Approaches,
  Applications, and Challenges.
\newblock \emph{ICT Express}, 10: 132--155.

\bibitem[{Reimers and Gurevych(2019)}]{reimers-2019-sentence-bert}
Reimers, N.; and Gurevych, I. 2019.
\newblock Sentence-BERT: Sentence Embeddings using Siamese BERT-Networks.
\newblock In \emph{Proc. of EMNLP}. ACL.

\bibitem[{Rocktaschel et~al.(2016)Rocktaschel, Grefenstette, Hermann, Kocisky,
  and Blunsom}]{rocktaschel2016reasoning}
Rocktaschel, T.; Grefenstette, E.; Hermann, K.~M.; Kocisky, T.; and Blunsom, P.
  2016.
\newblock Reasoning about Entailment with Neural Attention.
\newblock In \emph{Proc. of ICLR}.

\bibitem[{Williams, Nangia, and Bowman(2018)}]{N18-1101}
Williams, A.; Nangia, N.; and Bowman, S. 2018.
\newblock A Broad-Coverage Challenge Corpus for Sentence Understanding through
  Inference.
\newblock In \emph{Proc. of NAACL}, 1112--1122. ACL.

\bibitem[{Wotzlaw and Coote(2013)}]{DBLP:journals/corr/WotzlawC13}
Wotzlaw, A.; and Coote, R. 2013.
\newblock A Logic-based Approach for Recognizing Textual Entailment Supported
  by Ontological Background Knowledge.
\newblock \emph{CoRR}, abs/1310.4938.

\end{thebibliography}

\end{document}